\documentclass{article}
\newcommand{\headeright}{}
\usepackage{arxiv}
\newcommand{\undertitle}{}
\usepackage[utf8]{inputenc} 
\usepackage[T1]{fontenc}    
\usepackage{hyperref}       
\usepackage{url}            
\usepackage{booktabs}       
\usepackage{amsfonts}       
\usepackage{nicefrac}       
\usepackage{microtype}      
\usepackage{grap hicx}
\usepackage[numbers]{natbib}
\usepackage{doi}
\usepackage{amsmath}
\usepackage{float}

\title{Dynamic Lagging for Time-Series Forecasting in E-Commerce Finance: Mitigating Information Loss with A Hybrid ML Architecture}

\author{
\normalfont Abhishek Sharma$^{1}$, Anat Parush$^{1}$ \\
\normalfont Sumit Wadhwa$^{2}$, Amihai Savir$^{2}$ \\
\normalfont Anne Guinard$^{2}$, Prateek Srivastava$^{2}$ \\
\\
\normalfont Dell Technologies
\\ 
\\ 
}

\renewcommand{\shorttitle}{Dynamic Lagging for Time-Series Forecasting in eCommerce Finance}

\hypersetup{
pdftitle={Dynamic Lagging for Time-Series Forecasting in e-Commerce Finance: Mitigating Information Loss with a Hybrid ML Architecture},
pdfsubject={cs.LG},
pdfauthor={Abhishek Sharma, Anat Parush, Sumit Wadhwa, Amihai Savir, Anne Guinard, Prateek Srivastava},
pdfkeywords={Time-Series Analysis, Feature Engineering, Sparse Data Modeling, E-commerce Analytics, Ensemble Learning, Regression Analysis},
}

\begin{document}
\maketitle

\begin{abstract}
Accurate forecasting in the e-commerce finance domain is particularly challenging due to irregular invoice schedules, payment deferrals, and user-specific behavioral variability. These factors, combined with sparse datasets and short historical windows, limit the effectiveness of conventional time-series methods. While deep learning and Transformer-based models have shown promise in other domains, their performance deteriorates under partial observability and limited historical data. 

To address these challenges, we propose a hybrid forecasting framework that integrates dynamic lagged feature engineering and adaptive rolling-window representations with classical statistical models and ensemble learners. Our approach explicitly incorporates invoice-level behavioral modeling, structured lag of support data, and custom stability-aware loss functions, enabling robust forecasts in sparse and irregular financial settings. 

Empirical results demonstrate an approximate 5\% reduction in MAPE compared to baseline models, translating into substantial financial savings. Furthermore, the framework enhances forecast stability over quarterly horizons and strengthens feature–target correlation by capturing both short- and long-term patterns, leveraging user profile attributes, and simulating upcoming invoice behaviors. 

These findings underscore the value of combining structured lagging, invoice-level closure modeling, and behavioral insights to advance predictive accuracy in sparse financial time-series forecasting.
\end{abstract}


\section{Introduction}
Time-series forecasting has undergone significant advancements with the emergence of deep learning techniques and Transformer-based architectures, which have demonstrated strong performance in capturing long-term dependencies and complex temporal patterns. However, in scenarios where data are sparse, exhibit irregular sampling, or provide only a limited historical window, these highly parameterized models often fail to achieve consistent gains over traditional approaches. Empirical studies have shown that classical statistical techniques, such as SARIMA and Holt-Winters, along with tree-based ensemble learners, frequently outperform deep learning models in low-data regimes. These findings highlight, that in such contexts, the effectiveness of forecasting often depends more on data enrichment, feature engineering, and the incorporation of domain-specific temporal dynamics rather than increasing model complexity alone.

Accurate forecasting plays a critical role in multiple domains, including economics, finance, and operational decision-making. In the context of e-commerce finance, Treasury teams face challenges related to liquidity management and capital allocation, where forecast accuracy directly impacts interest expense reduction, investment optimization, and cashflow stability. However, invoice-level payment behaviors exhibit high temporal variability due to irregular billing cycles, partial payments, and deferrals, making it difficult for conventional deep learning models to capture underlying patterns.

To address these challenges, we propose a hybrid forecasting framework that combines feature-rich statistical modeling with ensemble learners. Our approach integrates the following components:

\begin{itemize}
    \item \textbf{Dynamic lagged feature engineering and adaptive rolling windows:} 
    These are used to capture short- and long-term temporal dependencies by constructing structured lag features across varying time horizons.

    \item \textbf{Simulating upcoming payments to correct for partial visibility:} 
    Incorporating upcoming invoice data for CatBoost ensemble regressor to predict invoice closure dates.

    \item \textbf{Dual-window modeling:} 
    We adopt a two-stage rolling-window strategy, where \textit{short-window regressors} model high-frequency fluctuations within the first four weeks of a quarter, while \textit{long-window models} capture smoother invoice behaviors for weeks five through thirteen also stabilizing long-horizon accuracy.
\end{itemize}

Furthermore, we address bias introduced by lagged dependent variables by adopting regression models that account for residual autocorrelation. We also propose a lag-weighted regression that applies adaptive penalties based on both coefficient size and lag influence, improving the balance between short- and long-term dependencies.

By leveraging structured lag of support data, invoice-level behavioral modeling, and custom loss functions, our framework advances prior work and provides a robust solution for forecasting under sparse, noisy, and partially observable conditions.

\section{Related Work}

Traditional time-series forecasting has historically relied on models such as \textbf{ARIMA}, \textbf{Holt-Winters}, and \textbf{exponential smoothing}. While these approaches provide reliable performance in stationary environments, they often struggle in dynamic financial settings characterized by \textbf{data sparsity}, \textbf{non-stationarity}, and \textbf{structural regime shifts}, such as quarter-end payment spikes.  
To address these challenges, rolling-window forecasting techniques have been widely studied. For example, Pesaran and Timmermann (2007) demonstrated that adaptive rolling window selection improves model stability and responsiveness to regime changes. Inspired by this, our approach segments data into \textbf{short-term} and \textbf{long-term} forecasting windows for each fiscal quarter to better handle varying payment behaviors.

Recent advancements in boosting frameworks have significantly improved forecasting performance for irregular financial datasets. Prokhorenkova et al. (2018) introduced \textbf{CatBoost}, a gradient-boosted decision tree framework highly effective for tabular forecasting due to its ability to natively handle categorical features, missing values, and non-uniformly distributed data. While gradient-boosted models have been successfully applied to problems such as \textbf{invoice payment prediction} and \textbf{anomaly detection}, prior works rarely integrate \textbf{dynamically engineered lagged features} across fiscal periods, limiting their ability to capture quarter-specific payment dynamics.

Similarly, Taylor and Letham (2017) introduced \textbf{Prophet}, a forecasting model designed to capture complex seasonal patterns and integrate external regressors in time-series forecasting tasks. However, Prophet’s application in enterprise-level \textbf{cash forecasting} has been limited, particularly in scenarios where \textbf{partial invoice observability} and \textbf{quarter-specific payment behaviors} significantly influence forecasting accuracy.

Recent findings from the M5 competition (Makridakis et al., 2022) further demonstrate that \textbf{hybrid forecasting architectures} combining statistical models and machine learning methods consistently outperform standalone approaches, especially in domains where seasonality, sparsity, and short historical windows coexist.  
Januschowski et al. (2022) provide a broader taxonomy of forecasting methodologies and emphasize the importance of leveraging dynamic regressors and hybrid techniques for improved performance.  

Our work builds on these insights and integrates them into a unified \textbf{hybrid forecast pipeline} tailored for enterprise finance. The proposed framework combines:
\begin{itemize}
    \item Structured lagging of support data to enhance correlation with the target variable,
    \item Invoice-level closure modeling using CatBoost,
    \item Multivariate time-series forecasting through Prophet,
    \item A custom stability-aware loss function designed to penalize fluctuations across the forecasting horizon.
\end{itemize}

This integration advances prior work by directly addressing challenges such as \textbf{partial observability}, \textbf{payment behavior variance}, and \textbf{forecast stability} across fiscal quarters, while aligning with recent trends in hybrid forecasting for enterprise-scale financial planning.

\section{Preliminaries}

\subsection{Baseline}
For comparative evaluation, we define the \textbf{baseline model} as a univariate forecasting architecture without:
\begin{enumerate}
    \item \textbf{Dynamic lag features} for support data, and
    \item \textbf{Rolling window horizons} for invoice-level modeling.
\end{enumerate}

The baseline forecast is computed by applying a simple univariate model on raw support data and aggregating the resulting forecasts at the \textbf{invoice level}. In this setup, future partial customer collections are \textbf{not modeled} explicitly, making it an appropriate benchmark for assessing the impact of \textbf{feature engineering} and \textbf{rolling horizon integration} in our proposed framework.

\subsection{Metric}

Traditional forecasting metrics such as \textbf{MAE}, \textbf{RMSE}, and \textbf{MAPE} measure pointwise prediction error but fail to account for \textbf{forecast stability} across rolling horizons. Since financial forecasting requires balancing \textbf{accuracy} and \textbf{variance control}, we propose a \textbf{variance-weighted evaluation metric}.

To emphasize the importance of \textbf{recent patterns} while maintaining robustness, we adopt a \textbf{3-fold sliding window} approach. We assign \textbf{increasing weights} to more recent folds, ensuring that:
\begin{enumerate}
    \item More recent observations contribute more strongly to the weighted error computation.
    \item The standard deviation of errors across folds is explicitly penalized to improve forecast stability.
\end{enumerate}

The variance-weighted evaluation metric is defined as:

\[
\text{Score} = \alpha \cdot \sum_{f=1}^{F} v_f \cdot \text{MAPE}_f
            + (1 - \alpha) \cdot \sqrt{ \sum_{f=1}^{F} v_f \cdot \left(\text{MAPE}_f - \overline{\text{MAPE}}\right)^2 },
\]

where:
\begin{itemize}
    \item $F = 3$ is the number of sliding folds,
    \item $\text{MAPE}_f$ = Mean Absolute Percentage Error for fold $f$,
    \item $\overline{\text{MAPE}} = \sum_{f=1}^{F} v_f \cdot \text{MAPE}_f$ = Weighted mean MAPE across folds,
    \item $v_f$ = Normalized weight assigned to fold $f$ such that $\sum_{f=1}^{F} v_f = 1$, with \textbf{larger $v_f$ for recent folds},
    \item $\alpha \in [0,1]$ = Trade-off parameter controlling emphasis on \textbf{accuracy} vs \textbf{stability}.
\end{itemize}

The final evaluation metric across all $k$ sliding windows within each fold is computed as:

\[
\text{Final Score} = \frac{1}{k} \sum_{i=1}^{k} \text{Score}_i,
\]

where $\text{Score}_i$ is the variance-weighted score for the $i$-th rolling window.

\subsection{Accuracy}
Accuracy is reported as the \textbf{uplift} of our proposed hybrid model relative to the univariate baseline, defined as:

\[
\text{Accuracy Uplift} = 
\frac{\text{Error}_{\text{baseline}} - \text{Error}_{\text{proposed}}}
{\text{Error}_{\text{baseline}}} \times 100,
\]

where $\text{Error}$ is computed using the variance-weighted score.  
This formulation directly quantifies the improvement achieved by integrating \textbf{dynamic lag features}, \textbf{Prophet decomposition based features}, and \textbf{rolling window modeling} into the forecasting pipeline.

\subsection{Invoice-Level Baseline}
To further evaluate model performance at the \textbf{invoice level}, we compare predicted payment closure dates against \textbf{ideal payment schedules} derived from customer-specific payment terms.  
For each invoice, we define the payment deviation as follows:

\[
\Delta t = \mathrm{Actual\ Payment\ Date} - \mathrm{Due\ Date},
\]

where a smaller $\Delta t$ indicates more accurate forecasting of invoice closures.  
This deviation-based evaluation complements aggregate metrics like \textbf{MAPE} and \textbf{RMSE}, providing fine-grained insights into \textbf{customer-level payment behaviors} and improving the interpretability of model outputs.

\section{User Profile Attributes}
User information empowers the model to adapt its behavior based on individual user profiles, allowing for personalized responses and predictions. We considered the following user profile attributes for incorporation into the proposed machine learning framework.

\subsection{Customer Definition}
We segment individual customers based on their \textbf{market segments} and \textbf{payment terms} to enable more accurate modeling of payment behaviors.  
A customer can belong to one of several primary segments:
\begin{itemize}
    \item \textbf{Consumer and Small Business (CSB)}
    \item \textbf{Commercial}
    \item \textbf{Enterprise}
\end{itemize}
Payment patterns vary significantly across these segments due to differences in billing policies, credit cycles, and contractual obligations.  
By incorporating these attributes, we ensure that the engineered payment features maintain \textbf{minimal intra-segment deviation} while capturing \textbf{inter-segment variability}, improving overall model accuracy.

\subsection{Delay in Payment}
For each customer, we compute a \textbf{speed-to-pay metric}, defined as the deviation between the \textit{actual payment date} and the \textit{invoice due date}.  
Formally:
\[
D_i = \text{Payment Date}_i - \text{Due Date}_i,
\]
where $D_i$ represents the delay (in days) for invoice $i$.  
To account for evolving payment behaviors, this metric is further adjusted using a \textbf{recency weight}, giving higher importance to more recent transactions.

\subsection{Average Payment}
The \textbf{average payment} made by a customer over a defined time horizon is computed as:
\[
\bar{P} = \frac{1}{N} \sum_{i=1}^{N} P_i,
\]
where $P_i$ represents the amount paid for invoice $i$, and $N$ denotes the total number of invoices.  
This measure captures a customer's typical payment magnitude and supports downstream segmentation.

\subsection{Deviation in Payment}
To measure payment variability, we calculate the \textbf{standard deviation} of payment amounts for each customer:
\[
\sigma_P = \sqrt{\frac{1}{N} \sum_{i=1}^N (P_i - \bar{P})^2}.
\]
Customers with higher $\sigma_P$ exhibit more irregular payment behaviors, which are particularly important for modeling invoice closure risk and forecast uncertainty.

\subsection{Recent Payment Behavior}
To emphasize the importance of temporal dynamics, we define a \textbf{recency-adjusted speed-to-pay metric} over the \textbf{90 days preceding the most recent transaction}.  
This captures short-term payment trends, allowing the model to adapt quickly to sudden behavioral changes, such as temporary liquidity constraints or improved payment cycles.

\section{Challenges and Motivation}

Forecasting customer payment behavior in the e-commerce finance domain presents several challenges due to the inherent variability across customer segments and the dynamic nature of financial transactions. Customers belong to distinct market segments such as \textbf{Consumer and Small Business (CSB)}, \textbf{Commercial}, and \textbf{Enterprise}, each exhibiting different payment distributions and behavioral patterns. While the commercial segment tends to demonstrate relatively stable payment cycles, CSB and enterprise customers display high variability in invoice closure timelines and payment delays. Designing an algorithm that can accurately \textbf{segment customers}, adaptively model their behavioral patterns, and handle \textbf{new customer onboarding} based on limited transaction history is a key challenge.

Another major complexity arises from the \textbf{real-time nature of invoice generation} and associated support data such as orders and collections. New invoices are continuously added to the system, requiring models to dynamically update forecasts while maintaining consistency across both \textbf{short-term} and \textbf{long-term} horizons. Traditional forecasting techniques often fail to account for these real-time fluctuations, leading to unstable predictions.

The motivation behind this study stems from the significant \textbf{financial implications} of improved forecasting accuracy. For enterprises handling \textbf{billions in yearly customer collections}, even a modest \textbf{1\% improvement in forecasting accuracy} can translate into \textbf{millions of dollars saved annually} through reduced interest expenses and improved liquidity management. Given the scale and dynamic nature of e-commerce operations, it becomes critical to design forecasting frameworks that:
\begin{itemize}
    \item Incorporate \textbf{dynamic customer payment behavior},
    \item Efficiently handle \textbf{real-time invoice inflows},
    \item Capture both \textbf{short-term trends} and \textbf{long-term seasonality}, and
    \item Enable more informed and adaptive \textbf{business decision-making}.
\end{itemize}

By addressing these challenges, we aim to develop a robust forecasting framework that integrates dynamic lagged feature engineering, adaptive rolling windows, and ensemble learning to improve accuracy and stability in financial forecasting.

\section{Hypothesis}

To design the formal algorithm and address the primary research questions, we investigate the following aspects:

\begin{enumerate}
    \item \textbf{Mitigating information loss:} Reducing the loss of information from open invoices by incorporating upcoming invoice data using a windowed forecasting approach, thereby improving customer collections predictions.
    \item \textbf{Enhancing feature-target correlation:} Improving the relationship between supporting regressors (e.g., order-level data, collections data) and the forecasted target variable.
    \item \textbf{Stabilizing long-horizon forecasts:} Ensuring stability of forecasts over extended horizons (e.g., quarterly for our case) while also minimizing week-over-week fluctuations.
\end{enumerate}

User profile attributes play a significant role in driving model performance. Based on this, we formulate two hypotheses:

\textbf{H1:} \textit{Baseline modeling using raw data.}  
The baseline model operates on support data without any dynamic lagged components, and invoice data is projected using a univariate time-series model. This setup reflects a simplified forecasting approach without feature enrichment.

\textbf{H2:} \textit{Dynamic modeling with enriched user-centric attributes.}  
In this hypothesis, invoice-level models incorporate user-specific attributes, allowing the framework to better represent individual customer behavior. A rolling-window mechanism is used to simulate upcoming invoices and project them onto historical records. Since support data, such as customer collections, typically arrives with lags, its distribution is modeled dynamically into lagged components. For example, payment behaviors observed in quarters Q1 to Q3 exhibit similar patterns, whereas Q4 follows a distinct distribution.

For both hypotheses, the forecasting architecture relies on a \textbf{multivariate time-series framework} optimized using a \textbf{custom loss function} weighted by the cross-validated mean and standard deviation of MAPE.  

\subsection*{Formalization of H1:}

Let $S$ represent the forecasting system, and let:

\[
L = \{l_1, l_2, l_3, \dots, l_m\} \quad \text{be the set of customer-specific attributes},
\]
\[
U = \{u_1, u_2, u_3, \dots, u_n\} \quad \text{be the set of generic invoice and order-level attributes},
\]
where $m, n \in \mathbb{N}$.

A boosting-based ensemble model uses $L$ and $U$ as features to infer the closing dates of open invoices by learning from historical customer payment patterns. The predictions are aggregated weekly, and a univariate time-series model is built on the aggregated data to project forecasts over a quarter. Support data for the same period is forecasted independently using univariate modeling, without introducing any lag-based mechanism.

The resulting forecasts, together with derived supporting regressors, are then provided to a \textbf{multivariate Prophet model}. However, in this setup, future missing open invoices are not accounted for, and autocorrelation in support data is ignored, as customer payments are often delayed relative to invoice generation.

\subsection*{Formalization of H2:}

For the second hypothesis, we simulate the partial availability of open invoice information by predicting closure dates and projecting this information backward to augment historical records. Support data is treated differently: its distribution is modeled at a quarterly level and transformed into \textbf{dynamically lagged features}, improving correlation with the target variable compared to using raw, unlagged support data.

This framework enables the model to leverage:
\begin{itemize}
    \item Customer-specific attributes for improved personalization,
    \item Rolling-window forecasting for simulating future invoice behavior,
    \item Dynamically lagged support data for stronger feature-target alignment.
\end{itemize}

By comparing H1 and H2, we aim to demonstrate the contribution of \textbf{dynamic lagged modeling} and \textbf{user-centric attributes} in improving forecast accuracy and stability.

\section{Methodology}

Our methodology focuses on building a \textbf{multivariate forecasting framework} using partial open invoice information, designed to improve forecast stability and accuracy under sparse and dynamic conditions. A \textbf{rolling-window approach} is used to simulate the current state of invoices and project this behavior onto historical data, thereby enabling the model to account for evolving customer payment patterns.

To handle variability across different quarters, the dataset is segmented based on the \textbf{payment distribution} of supporting data sources, and \textbf{dynamically lagged regressors} are engineered prior to final modeling. This ensures that a single model does not attempt to learn heterogeneous behaviors across all periods, which would otherwise lead to suboptimal performance given the significant distributional shifts observed between quarters.

The total forecasted output represents \textbf{customer collections}, derived from two primary feature groups:  
(1) invoice-level attributes and  
(2) supporting data attributes.  
Empirically, invoice-level data demonstrates a strong linear relationship with the target, with correlations exceeding \textbf{70\%}, while for supporting data it is around \textbf{50\%}.

Unlike typical models that implicitly assign uniform weight to all features, our framework explicitly models these relationships using the \textbf{Prophet model}, chosen for its \textbf{interpretability} and ability to incorporate \textbf{external regressors}. Supporting features include historical and forecasted collections derived from invoices, simulated using the earlier rolling-window approach. Prophet’s interpretability allows us to evaluate the contribution of supporting attributes, seasonal components, and external effects to overall forecast accuracy.

Additionally, holiday effects such as \textbf{Black Friday sales} and global disruptions such as \textbf{COVID-19} are integrated as custom regressors to better capture domain-specific anomalies. To handle recurring seasonal effects, we use Fourier transformations within Prophet to model \textbf{weekly}, \textbf{quarterly}, and \textbf{yearly} seasonality. These Fourier coefficients are tuned using \textbf{Bayesian optimization}, enabling the model to adapt to seasonal irregularities — particularly in \textbf{Q4}, where the distribution diverges significantly from the rest of the year.

We further analyze the model’s component-wise breakdown to understand the contribution of each signal to final forecasts. While external regressors exhibit high correlation with the target, their direct contribution to predictions ranges between \textbf{10\% and 25\%}. This reduced impact arises because the target variable, \textbf{Total Customer Collections}, has lower variance, and Prophet’s piecewise linear model prioritizes the overall growth term over localized fluctuations.

\subsection*{Summary of Key Components}
Our forecasting framework integrates the following components:

\begin{enumerate}
    \item \textbf{Invoice-Level Modeling:} Using CatBoost to predict invoice closure dates, capturing invoice-specific patterns effectively with corrections to augment partial visibility.
    \item \textbf{Rolling Window Simulation:} Employing both \textit{short-term windows} (4 weeks) to capture near-term dynamics and \textit{long-term windows} (13 weeks) for quarterly trends.
    \item \textbf{Dynamic Lagged Regressors:} Generating quarter-specific lagged features to improve the correlation between supporting data and the target.
    \item \textbf{Multivariate Prophet Integration:} Combining forecasts and supporting regressors within a Prophet framework enhanced with external features, holiday effects, and a custom loss function optimized for stability and accuracy.
\end{enumerate}

\subsection{Invoice-Level Modeling}

We construct \textbf{customer-based features} using data from daily invoices to model payment behavior and predict invoice closure timelines. The \textbf{CatBoost regression model} is employed to estimate invoice closing dates by leveraging a combination of:
\begin{itemize}
    \item \textbf{Engineered features}, such as historical customer payment patterns, payment delays, and prior collection trends.
    \item \textbf{Raw invoice attributes}, including invoice amount, expected closing date, default payment indicators, and payment terms.
\end{itemize}

Once invoice-level closing dates are predicted, we aggregate weekly collections and use them as inputs for a \textbf{univariate forecasting model}. To effectively capture temporal dependencies, we adopt a \textbf{rolling-window mechanism} using both short-term and long-term regressors. Specifically, the framework predicts customer collections at a given time point based on historical aggregated windows of size $x$ (short-term) or $y$ (long-term), thereby retaining information from \textbf{partially open invoices}. These “partial windows” are then provided as inputs to a \textbf{multivariate Prophet time-series model} for forecasting upcoming invoices.

\subsubsection*{Rolling Window Illustration}

\begin{figure}[h]
\centering
\includegraphics[width=0.8\linewidth]{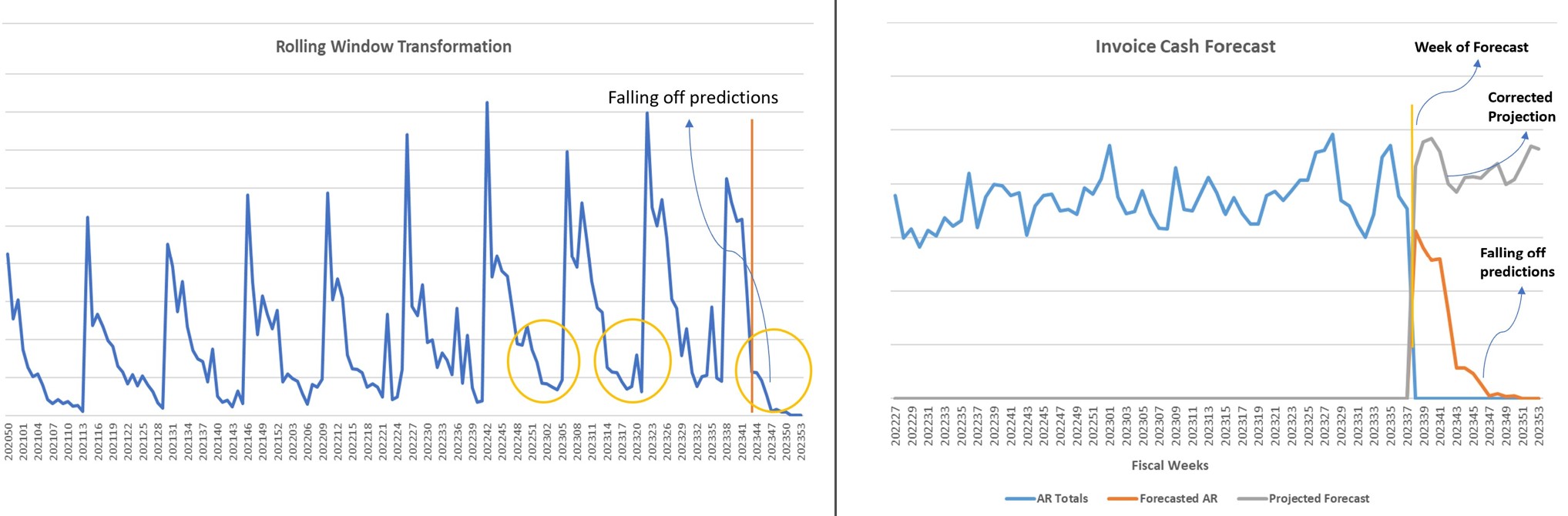}
\caption{Lags per quarter}
\label{fig:lags_per_quarter}
\end{figure}

Figure~\ref{fig:lags_per_quarter} illustrates the quarterly lag structure using both a 4-week (short-term) and a 13-week (long-term) window. In week 37, for example, several invoices remain open. Since there is limited visibility of invoices during quarter-end, aggregating historical collections and combining them with the CatBoost model enables accurate estimation of \textbf{invoice closure dates}. Collections per week are aggregated, and invoices projected to close in the current week are included, which explains the natural “tail-off” in the plotted curve.

The “\textbf{Forecasted AR}” in Figure~\ref{fig:lags_per_quarter} refers to aggregated customer collections computed from historical invoices using a \textbf{univariate time-series model}. For instance, at the start of a new quarter (e.g., week 39), we analyze historical behavior by rolling back four weeks to week 35, then further to week 31, and so forth. However, traditional univariate forecasts discard incoming invoice information, whereas our rolling-window mechanism explicitly integrates \textbf{partial invoice visibility} into the predictions.

\subsubsection*{Optimization of Short- and Long-Term Regressors}

To improve forecast accuracy, we train separate models with varying window lengths:
\begin{itemize}
    \item A short-term model using an $x$-week window (e.g., 4 weeks),
    \item A long-term model using a $y$-week window (e.g., 13 weeks, one quarter).
\end{itemize}

Each model is optimized independently to align predicted patterns with actual historical trends. We further integrate both short-term and long-term regressors into a combined modeling framework that fuses:
\begin{enumerate}
    \item \textbf{CatBoost-based invoice closure predictions}, and
    \item \textbf{Prophet-based multivariate forecast}.
\end{enumerate}

Finally, we employ \textbf{Bayesian optimization} to tune hyperparameters and optimize the short- and long-term forecasting windows, ensuring improved alignment between predicted and observed invoice closure behaviors over both granular and quarterly horizons.

\subsection{Support Data}

Support data serves as an important set of \textbf{external regressors} in our forecasting framework, complementing invoice-level attributes to improve the prediction of total customer collections (\textbf{Total Forecast}). These regressors include auxiliary operational signals, such as order-level activity, shipment details, and related behavioral indicators that correlate with future collections.

\begin{figure}[h]
\centering
\includegraphics[width=0.6\linewidth]{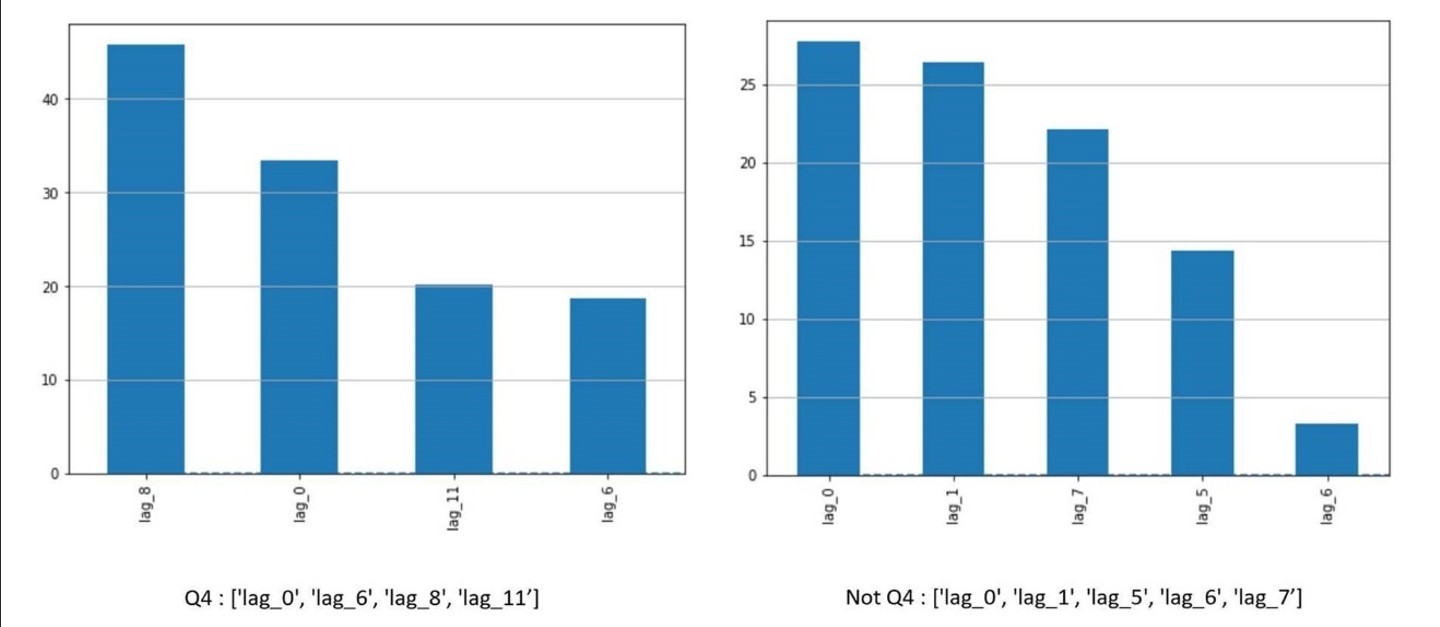}
\caption{Lag relevance for Q4 vs. non-Q4 periods}
\label{fig:lags_q4}
\end{figure}

To maximize the predictive power of these regressors, we transform raw support data into \textbf{lagged features}. Relevant lags are identified using a \textbf{linear regression model}, where we evaluate the contribution of each lag to explaining variance in total collections. Since customer payment behavior demonstrates significant seasonal variation, particularly in \textbf{Q4} due to fiscal year-end effects, we select lags separately for \textbf{Q4} and \textbf{non-Q4 periods}.  

Figure~\ref{fig:lags_q4} illustrates the difference in the relevant lag selections. This quarter-specific lag selection ensures that seasonality is better captured and improves the alignment between the regressors and the target.

Next, we aggregate customer collections derived from these lagged support data points to construct dynamic regressors tailored to each quarter.

\begin{figure}[h]
\centering
\includegraphics[width=0.8\linewidth]{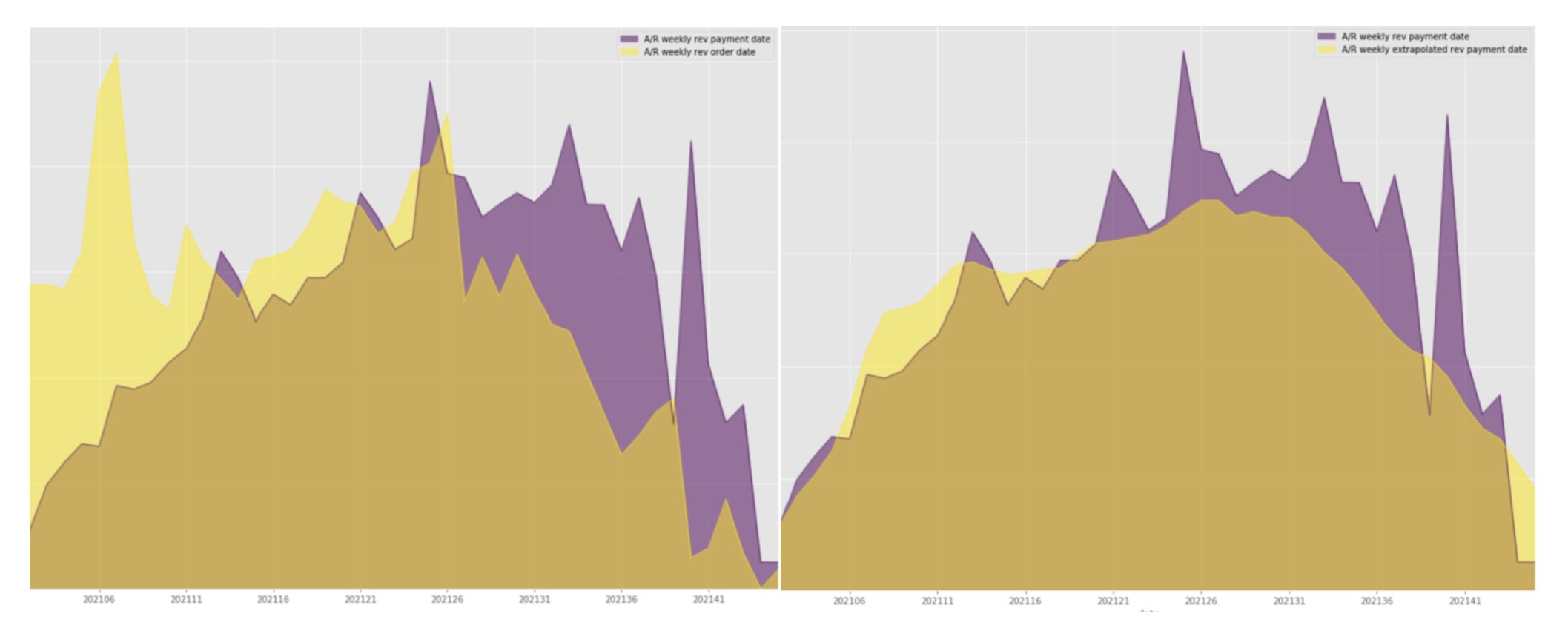}
\caption{Dynamic lag allocation strategy}
\label{fig:dynamic_lags}
\end{figure}

Figure~\ref{fig:dynamic_lags} demonstrates the effectiveness of the \textbf{dynamic lag allocation strategy}. By selecting quarter-specific relevant lags, the correlation between support data and customer collections improves substantially. As the forecasts generated using dynamically lagged support data overlap closely with actual observed data, this approach enhances the accuracy and stability of the final predictions.

\subsection{Overall Framework}

\begin{figure}[h]
\centering
\includegraphics[width=0.8\linewidth]{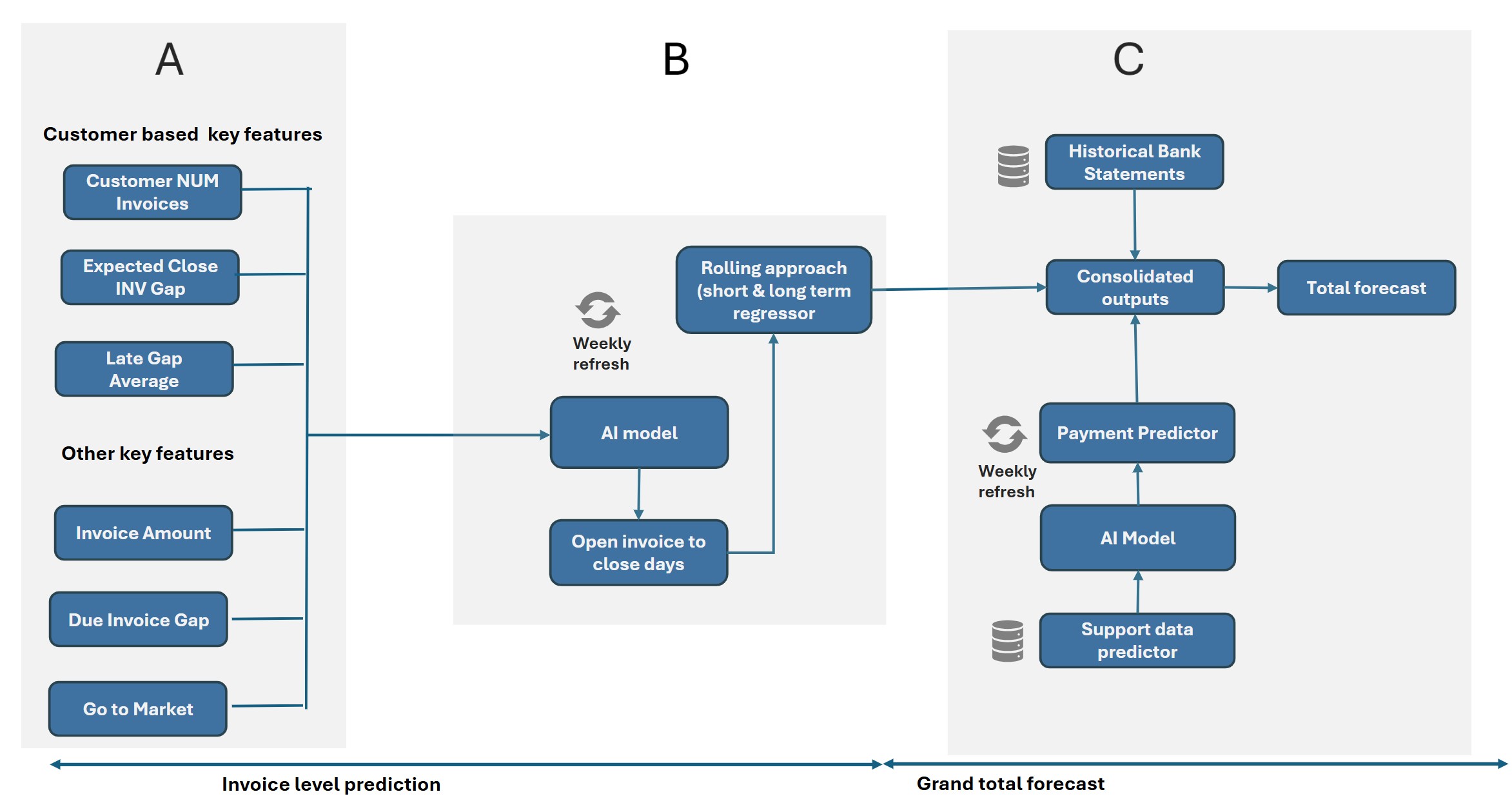}
\caption{Model Architecture}
\label{fig:model_architecture}
\end{figure}

Figure~\ref{fig:model_architecture} illustrates the overall \textbf{forecasting framework} combining \textbf{invoice-level modeling}, \textbf{support data regressors}, and \textbf{multivariate time-series forecasting}.  
The architecture begins with a \textbf{CatBoost regression model} trained on historical invoice data to predict closure dates of open invoices. The predicted closure dates and aggregated collections are then used to construct short-term and long-term \textbf{rolling-window regressors} (e.g., 4-week and 13-week windows), ensuring that partial information from invoices is retained.

In parallel, \textbf{supporting data sources} (e.g., orders, shipments, and auxiliary operational features) are processed using a \textbf{dynamic lag allocation strategy}, where quarter-specific relevant lags are selected to maximize correlation with the target variable (total customer collections). These engineered regressors complement invoice-level forecasts by incorporating seasonality and behavioral patterns derived from historical support data.

Finally, all components — predicted invoice closures, aggregated rolling-window collections, dynamically lagged support features, and external regressors (e.g., holiday effects and special events) — are integrated into a \textbf{multivariate Prophet model}. This unified architecture enables the framework to simultaneously capture:
\begin{itemize}
    \item \textbf{Short-term fluctuations} using 4-week rolling windows,
    \item \textbf{Long-term quarterly trends} via 13-week aggregated regressors,
    \item \textbf{Seasonality and special events} through Fourier-based components optimized via Bayesian tuning.
\end{itemize}

By combining these elements, the architecture improves overall forecast accuracy, ensures temporal stability, and adapts to the dynamic nature of invoice generation and customer payment behaviors.

\subsection{Custom Loss Function}

To improve predictive performance, we introduce a strategy that leverages \textbf{lagged variables} of both the target and selected features. Historical observations from previous time steps are used as predictors, but only the \textbf{most relevant lagged features}—those exhibiting strong correlation with the current target variable—are retained to optimize model accuracy.

In particular, we use \textbf{lagged order information} to improve the correlation between customer collections derived from orders and the overall total forecast. This incorporation of lagged regressors allows the model to better capture temporal dependencies within customer payment behavior.

\subsubsection*{Loss Function Formulation}

The custom loss function $L$ jointly optimizes for \textbf{average predictive accuracy} and \textbf{forecast stability} by combining the \textit{weighted mean of errors} and the \textit{weighted standard deviation of errors}:

\[
L = \alpha \cdot \overline{E}_w + (1 - \alpha) \cdot \sigma_w
\]

where:
\begin{itemize}
    \item $\alpha \in [0,1]$ is a hyperparameter controlling the trade-off between overall accuracy and forecast stability,
    \item $\overline{E}_w$ is the \textbf{weighted mean error across folds},
    \item $\sigma_w$ is the \textbf{weighted standard deviation of errors across folds}.
\end{itemize}

The weighted mean of cross-validated errors is defined as:

\[
\overline{E}_w = \frac{\sum_{k=1}^{K} w_k \, e_k}{\sum_{k=1}^{K} w_k}
\]

and the weighted standard deviation of errors across folds is computed as:

\[
\sigma_w = \sqrt{\frac{\sum_{k=1}^{K} w_k \, (e_k - \overline{E}_w)^2}{\sum_{k=1}^{K} w_k}}
\]

where:
\begin{itemize}
    \item $e_k$ represents the validation error (e.g., MAPE) for fold $k$,
    \item $w_k$ is the weight assigned to fold $k$ (e.g., based on recency or importance),
    \item $K$ is the total number of cross-validation folds.
\end{itemize}

\subsubsection*{Sliding Window Weighting Strategy}

To emphasize the importance of \textbf{recent patterns} while maintaining robustness, we adopt a \textbf{3-fold sliding window} approach. We assign \textbf{increasing weights} to recent folds, ensuring that:
\begin{enumerate}
    \item More recent observations contribute more strongly to the weighted error computation.
    \item The standard deviation of errors across folds is explicitly penalized to improve forecast stability.
\end{enumerate}

This weighted sliding-window strategy allows the model to adapt more effectively to dynamic invoice flows while reducing sensitivity to older, less relevant patterns.

During Bayesian optimization for the final ensemble, both $\alpha$ and the fold-level weights $\{w_k\}$ are tuned jointly to achieve an optimal balance between accuracy and consistency.

\section{Experimental Results and Discussion}

The primary contribution of this work lies in the development of a \textbf{machine-learning-driven forecasting framework} that effectively integrates \textbf{dynamic lagged features}, \textbf{rolling-window simulations}, and \textbf{multivariate time-series modeling} to improve forecast accuracy and stability.  
A key strength of the proposed system is its ability to seamlessly incorporate newly generated invoices into the forecasting pipeline, thereby retaining relevant information and adapting to changing payment behaviors.  
Additionally, the inclusion of dynamically engineered lags enables the model to better capture temporal dependencies between support data and realized customer collections.

\subsection*{Implementation Details}

We implemented our methodology on customer invoice datasets spanning \textbf{Commercial}, \textbf{Consumer and Small Business (CSB)}, and \textbf{Enterprise} segments. The implementation pipeline consisted of the following steps:

\begin{itemize}
    \item \textbf{Customer segmentation:} Segregated invoice data into distinct market segments and applied customer definitions to account for heterogeneity in payment patterns.
    \item \textbf{Feature aggregation:} Combined relevant invoice-level and support-level features, ensuring minimal variance in aggregated representations.
    \item \textbf{Invoice closure prediction:} Built a CatBoost regression model to predict invoice closure dates by learning from historically closed invoices.
    \item \textbf{Rolling-window simulations:} Aggregated weekly collections and applied short- and long-term rolling windows (4-week and 13-week) to better model temporal dependencies.
    \item \textbf{Dynamic lag engineering:} Applied dynamically selected lagged features on support data to maximize correlation with the target variable (total customer collections).
    \item \textbf{Multivariate forecasting:} Integrated all engineered features into a multivariate \textbf{Prophet} model for time-series-based forecasting.
    \item \textbf{Custom stability-aware loss:} Designed a custom loss function that penalizes excessive fluctuations in forecasts across the prediction horizon.
\end{itemize}

\subsection*{Advantages of the Proposed Approach}

Our experimental findings highlight several benefits of the proposed system:

\begin{enumerate}
    \item \textbf{Retention of partial invoice information:} Unlike traditional univariate forecasting, our approach effectively incorporates partially available invoice data into short- and long-term projections.
    \item \textbf{Adaptive retraining for directional accuracy:} Each retraining cycle updates the forecasts based on the latest invoice and support data, allowing the model to capture shifts in payment behaviors and trends.
    \item \textbf{Improved correlation with support data:} The introduction of \textbf{dynamic lagged regressors} significantly enhances the correlation between auxiliary operational features and the target variable, improving overall predictive performance.
\end{enumerate}

\section{Conclusion}

In this paper, we proposed a \textbf{novel forecasting architecture} that integrates \textbf{dynamic lagged feature engineering}, \textbf{rolling-window simulations}, and \textbf{multivariate time-series modeling} to improve the accuracy and stability of customer collection forecasts over extended horizons. Our approach effectively addresses challenges related to \textbf{tailing-off values}, \textbf{partial invoice visibility}, and \textbf{temporal variability} in payment behaviors.

Through extensive experimentation, we observed an approximate \textbf{5\% improvement in MAPE} over a baseline univariate forecasting setup, measured over one year of production monitoring. The inclusion of dynamic lagged regressors, combined with short-term and long-term modeling strategies, significantly improved the correlation between support data and the target variable while ensuring stable performance across quarterly horizons.

Furthermore, we explored the use of a \textbf{variable sliding weighted window} that assigns greater importance to recent observations. While this approach effectively modeled recency effects, its performance was comparable to the simpler fixed-window implementation. Therefore, for interpretability and efficiency, we adopted the simpler strategy in the final framework.

In general, our work demonstrates that combining \textbf{structured invoice modeling}, \textbf{dynamically engineered lagged features}, and \textbf{custom stability-aware optimization} results in a robust forecasting framework capable of adapting to evolving customer behaviors and improving long-term financial planning outcomes.

\end{document}